\pdfoutput=1

\documentclass[11pt]{article}

\usepackage{naacl2021} 

\usepackage{times}
\usepackage{latexsym}
\usepackage{framed}
\usepackage{amsmath} 
\usepackage{graphicx}
\usepackage{dsfont}
\usepackage{subfigure}
\usepackage{pdfcomment}
\usepackage{pifont}
\usepackage{enumitem}

\usepackage[T1]{fontenc}

\usepackage[utf8]{inputenc}

\usepackage{microtype}

\title{Instructions for NAACL-HLT 2021 Proceedings}

\title{News Headline Grouping as a Challenging NLU Task}

\author{Philippe Laban \\
  UC Berkeley \\\And
  Lucas Bandarkar \\
  UC Berkeley \\\And
  Marti A. Hearst\thanks{~Author emails: \{phillab, lucasbandarkar,hearst\}@berkeley.edu}\\
  UC Berkeley\\
  }

\date{}

\begin{document}
\maketitle
\begin{abstract}
    Recent progress in Natural Language Understanding (NLU) has seen the latest models outperform human performance on many standard tasks. These impressive results have led the community to introspect on dataset limitations, and iterate on more nuanced challenges.
    In this paper, we introduce the task of HeadLine Grouping (HLG) and a corresponding dataset (HLGD) consisting of 20,056 pairs of news headlines, each labeled with a binary judgement as to whether the pair belongs within the same group. 
    On HLGD, human annotators achieve high performance of around 0.9 F-1, while current state-of-the art Transformer models only reach 0.75 F-1, opening the path for further improvements.
    We further propose a novel unsupervised Headline Generator Swap model for the task of HeadLine Grouping that achieves within 3 F-1 of the best supervised model.
    Finally, we analyze high-performing models with consistency tests, and find that models are not consistent in their predictions, revealing modeling limits of current architectures.
\end{abstract}

\section{Introduction}

Headlines are a key component in everyday news consumption. As the first piece of text the user interacts with when learning about a story, the headline can play many roles, including: summarize the main points of the story, promote a particular detail, and convince the reader to choose one source over another \cite{bonyadi2013headlines}.

News aggregators amass content from many disparate news sources and have become popular, in part because they offer news readers access to diverse sources \cite{chowdhury2006news}. \citet{flaxman2016filter} find that news aggregators help news readers access content they are unfamiliar with, and potentially on opposite sides of the political spectrum. At the heart of a news aggregator is the ability to group relevant content together, to support a reader in finding varying views and angles on the news.

Natural Language Understanding (NLU) has seen rapid progress in recent years. The creation of multi-task benchmarks such as the General Language Understanding Evaluation collection (GLUE), paired with fast-paced progress in Transformer-based architectures has led to models outperforming human baseline performance on many tasks, such as paraphrase identification \cite{dolan2004mrpc}, semantic similarity \cite{cer-etal-2017-semeval}, and extractive question-answering (QA) \cite{rajpurkar-etal-2018-know}.

This success has led to the questioning of the composition of benchmarks, and the subsequent creation of ever-more challenging datasets, for example by increasing the diversity of texts in textual entailment datasets \cite{williams-etal-2018-broad}, or introducing unanswerable questions in QA datasets \cite{rajpurkar-etal-2018-know}.

\begin{figure}
    \centering
    \setlength{\belowcaptionskip}{-12pt} 
    \includegraphics[width=0.47\textwidth]{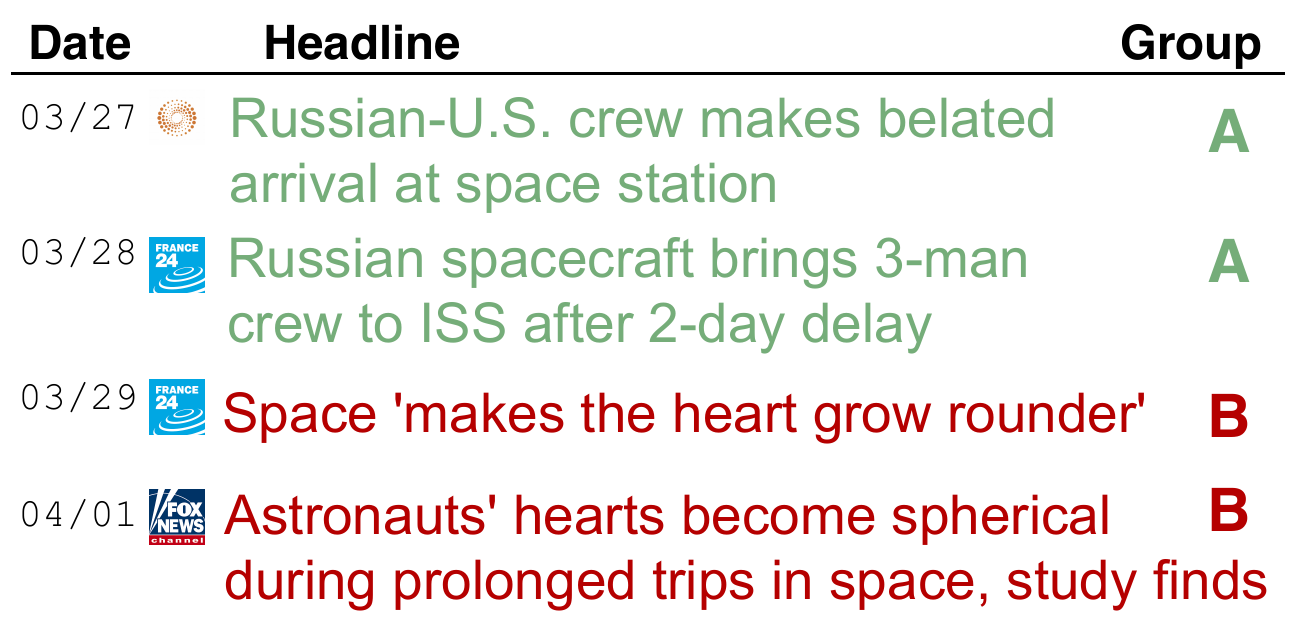}
    \caption{\textbf{Snippet of timeline in the HeadLine Grouping dataset (HLGD).} The headlines are part of the Space timeline, one of 10 timelines in HLGD. Headlines labeled A are part of a group; those labeled B are part of another group.}
    \label{fig:hp_example}
\end{figure}

\subsection{HeadLine Grouping Definition}

In this paper, we propose the novel task of HeadLine Grouping.
Although news articles may discuss several topics, because of length constraints, headlines predominantly describe a single event. Therefore, for the task of headline grouping, we define two headlines to be in the same group if they describe the same event: an action that occurred at a specific time and place. We do not require headlines to contain fully identical information to be placed into the same group. For example, one headline might report an exact number of deaths, while another might report a rounded number, or omit the number altogether.
Figure~\ref{fig:hp_example} shows an example from our dataset. The first two headlines are in group A, and the third and fourth are part of group B. The headlines are divided into groups A versus B because they describe different events in the timeline (astronauts arriving at the space station vs. a study about hearts in space).
The two headlines in B show the lexical and syntactic diversity of groups in this dataset -- they appear in the same group because they describe the same underlying event. Appendix~\ref{appendix:longer_excerpt} gives a longer excerpt.

We build a large dataset for the task of HeadLine Grouping, crowd-sourcing the annotation of large timelines of news headlines in English. We cast the task as a binary classification: given a pair of headlines, determine whether they are part of a headline group (1) or whether they relate to distinct events (0).

\subsection{Contributions}

Our main contribution, described in Section~\ref{section:dataset}, is the design of the HeadLine Grouping task (HLG), and the creation of the HeadLine Grouping Dataset (HLGD) that is focused on detecting when headlines refer to the same underlying event. We show that the human annotations in our dataset have strong inter-annotator agreement (average 0.81), and a human annotator can achieve high performance on our corpus (around 0.9 F-1), while current state-of-the-art Transformer-based model performance stands around 0.75 F-1.

A second contribution is a novel unsupervised approach for the task of HeadLine Grouping relying on modifying a headline generator model. The model achieves the best performance on HLGD amongst unsupervised methods. Section~\ref{section:results} presents the performance of this algorithm compared to several baselines, including supervised and unsupervised methods.

Our final contribution, presented in Section~\ref{section:analysis}, is an analysis of the consistency of the best performing model on HLGD. We specifically analyze whether the model follows commutative and transitive behavior expected to be trivially true in HeadLine Grouping.\footnote{The code, model checkpoints and dataset are available at: \url{https://github.com/tingofurro/headline_grouping}}

\section{Related Work}

\textbf{Paraphrase Identification, Textual Entailment and Semantic Similarity} are three common NLP tasks that resemble HeadLine Grouping.
In Paraphrase Identification (PI) \cite{eyecioglu2016paraphrase,Xu2014ExtractingLD}, the objective is to determine whether two sentences are semantically equivalent. We show in Table~\ref{table:positive_negative} that only one third of positive headline pairs in HLGD qualify as paraphrases. We further show in Section~\ref{section:results} that a trained model on MRPC \cite{dolan2004mrpc}, a PI dataset of news text, performs poorly on HLGD.
Textual entailment \cite{Bentivogli2009TheSP}, or Natural Language Inference (NLI) \cite{williams-etal-2018-broad}, determines whether a premise text implies a hypothesis. Apart from the non-symmetricality of the entailment relationship, we believe entailment is not well-suited to the domain of headlines because of the strict nature of the relationship. A large portion of headlines in a group differ in level of detail, and under an entailment task, would need to be labeled as neutral or contradicting.
Finally, semantic similarity assigns a strength of similarity between two candidate sentences, for example in the Semantic Textual Similarity Benchmark (STS-B) \cite{cer-etal-2017-semeval}, similarity is ranked from 1 to 5. This flexibility seems like a good fit; however, the lexical and syntactic diversity of headlines about the same underlying content do not correspond well to a similarity range.

\textbf{Topic Detection and Tracking} (TDT) \cite{allan2002introduction} was a DARPA-sponsored initiative to investigate methods to group news articles by topics (referred to as timelines in this paper). We view TDT as a precursor to the task of HeadLine Grouping: in TDT, the focus is on detecting and tracking a timeline of related events, while in HeadLine Grouping, the timeline is given, and the focus is on subdividing it into finer groups.
We considered using the TDT datasets and annotating them for our purposes. However, the TDT developers acknowledge \cite{Graff2006TDT5} several important errors in the way the TDT datasets were acquired (e.g., some publication dates were not properly attributed) that could have an impact on the quality of the final dataset.

\textbf{News Headlines in NLP.} Headlines are popular as a challenging source for generation tasks such as summarization \cite{rush-etal-2015-neural}, style transfer \cite{jin-etal-2020-hooks}, and style-preserving translation \cite{Joshi2013MakingHI}. Headlines have been leveraged to detect political bias \cite{Gangula2019DetectingPB}, click-bait and fake news phenomena \cite{Bourgonje2017FromCT}. Finally, sentiment analysis of headlines has received attention \cite{bostan-etal-2020-goodnewseveryone}, with some work showing headline sentiment can be a useful signal in finance \cite{moore-rayson-2017-lancaster}.

\textbf{Grouping Headlines} has been explored in prior work. \citet{wubben2009clustering} propose a TF-IDF based clustering algorithm, but do not evaluate its agreement with human annotations. \citet{pronoza2015construction} build a corpus of Russian headlines pairs, but limit pairs in the dataset by filtering out headlines that are distant syntactically. We find that 
headline groups often contain syntactically distant headlines (see Figure~\ref{fig:levenshtein_distance}).
\citet{bouamor2012etude} and \citet{shinyama2002automatic} present a simple strategy, relying on the assumption that all articles on a topic published on the same day form a group. As will be shown below, this assumption is not always correct (see Figure~\ref{fig:time_analysis}).

Several of the most-used news aggregators, such as Yahoo News \footnote{https://news.yahoo.com}, Google News\footnote{https://news.google.com}, and Bloomberg's NSTM \cite{bambrick2020nstm} present headlines in groups. As these systems do not have published algorithms, we cannot comment on their methods; nonetheless we hope that the release of the HLG dataset offers a common evaluation test-bed to benchmark systems.

\section{HeadLine Grouping Dataset}
\label{section:dataset}

We now present the HeadLine Grouping Dataset. We describe the dataset of news articles we collected for annotation, our annotation procedure, an analysis of the resulting dataset, and the challenges we propose to the community.

\subsection{Dataset Source}

\begin{table}[]
    \resizebox{0.48\textwidth}{!}{%
    \begin{tabular}{lcccc}
    \hline
    \textbf{Story Name}         & \textbf{Size} & \textbf{Groups} & \textbf{+ pairs} & \multicolumn{1}{l}{\textbf{IAA}} \\ \hline
    Tunisia Protests            & 111           & 46              & 219              & 0.758                            \\
    Ireland Abortion Vote & 180           & 81              & 406              & 0.727                            \\
    Ivory Coast Army Mutiny     & 128           & 45              & 329              & 0.781                            \\
    International Space Station & 257           & 107             & 499              & 0.831                            \\
    US Bird Flu Outbreak        & 79            & 36              & 91               & 0.924                            \\
    Human Cloning               & 119           & 55              & 259              & 0.830                            \\ \hline
    Facebook Privacy Scandal    & 194           & 105             & 274              & 0.753                            \\
    Equifax Breach              & 159           & 81              & 261              & 0.855                            \\ \hline
    Brazil Dam Disaster         & 273           & 132             & 634              & 0.818                            \\
    Wikileaks Trials            & 180           & 101             & 550              & 0.859                            \\ \hline
    \textbf{Total / Average}          & \textbf{1679} & \textbf{789}    & \textbf{3522}    & \textbf{0.814}                   \\ \hline
    \end{tabular}
    }
    \caption{\textbf{Names and statistics of the ten news timelines in HLGD}. \textit{Size} is the number of headlines in the timeline, \textit{Groups} the number of distinct headline groups, \textit{+ pairs} the number of pairs of headlines in all groups, and \textit{IAA} the inter-annotator agreement. Timelines are separated into training (1-6), development (7-8), and test (9-10).}
    \label{table:data_source}
\end{table}

We collect a set of 10 news timelines from an existing open-source news collection in English \cite{laban2017newslens}. A timeline is a collection of news articles about an evolving topic, consisting of a series of events. The timelines we use to build HLGD consist of time-stamped English news articles originating from 34 international news sources. The timelines range in size from 80 to 274 news articles, and span 18 days to 10 years.

We choose to use timelines as the source for the dataset for two reasons. First, news timelines center around a theme, and as successive events occur, many pairs of headlines will be semantically close, yielding challenging samples for the dataset. Second, this task requires annotating headlines by pairs. If there are $n$ headlines, there could be on the order of $n^2$ headline pairs to annotate. By having annotators assign group labels to a chronologically organized timeline, the annotation procedure requires only one label per headline, or $n$ labels total.

We attempted to diversify topics and geographical locations represented in the 10 selected timelines. Topics and statistics are shown in Table~\ref{table:data_source}. 

\subsection{Annotation Procedure}

To reduce the effects of varying judgement inherent to the task, annotations were obtained from five independent judges and merged using the procedure described in the following subsection. Annotators worked on an entire timeline at a time, using the following  procedure:

\begin{itemize}
\setlength\itemsep{-0.1em}
    \item The timeline was presented in a spreadsheet, in chronological order, with a single headline per row, and the corresponding publication date (year, month, day),
    \item Annotators went over the timeline one headline at a time in chronological order,
    \item If the headline being annotated did not match a previously created group, the annotator assigned it a new \textit{group number},
    \item Otherwise, the annotator could assign the headline to a previous \textit{group number}, grouping it with previously added headlines.
\end{itemize}

We note that the annotation relied on annotators' ability to discern an event described by a news headline.  However, a headline is not always written in an event-centric manner -- for example when the headline is vague (e.g., \textit{A way forward in gene editing} in the Human Cloning timeline), or overly brief (e.g., \textit{Waste not, want not} in the International Space Station timeline). Annotators were instructed to create a separate group for such cases, isolating non-event-centric headlines.

Roughly one fifth of the annotations were produced by authors of the paper, and the remaining annotations were obtained by recruiting 8 crowd-workers on the Upwork platform.\footnote{https://www.upwork.com} The crowd-workers were all native English speakers with experience in either proof-reading or data-entry, and were remunerated at \$14/hour.

Annotators were first trained by reading a previously annotated timeline, and given the opportunity to clarify the task before starting to annotate. Exact instructions given to the annotators are transcribed in Appendix~\ref{appendix:annotator_intructions}.

\subsection{Merging Annotations}

In order to merge the five annotations, we follow a standard procedure to produce a single grouping that represents an aggregate of annotations.

We create a graph $G$, with each headline in a timeline represented by a node $n_i$. An edge $(n_i,n_j)$ is added to $G$ if a majority of the annotators (three or more of the five) put the two headlines in the same group. We apply a community detection algorithm, the Louvain method \cite{blondel2008fast}, to $G$ to obtain a grouping of the headlines that we call the \textit{global groups}.

\subsection{Inter-annotator Agreement}
\label{sec:inter_annotator_agreement}

We compare the groups of each annotator to the global groups for each timeline, and measure agreement between annotator groups and a leave-one-out version of the \textit{global groups} using the standard Adjusted Mutual Information \cite{vinh2010information}. The average inter-annotator agreement is 0.814, confirming that consensus amongst annotators is high. Inter-annotator agreement is reported for each timeline in Table~\ref{table:data_source}.

Section~\ref{section:results} provides individual annotator performance on HLGD, which obtain the highest performance of about 0.9 F-1, further confirming that the task is well defined for human annotators.

\subsection{Creating the Final Dataset}

\begin{table*}[]
    \centering
    \resizebox{0.98\textwidth}{!}{%
    \def\arraystretch{1.1}
	\begin{tabular}{|p{2cm}p{5cm}p{10cm}r|}
	\hline
	\multicolumn{4}{|c|}{\textbf{Positive Examples}} \\
	Reasoning & Description & Example Headline Pair & Percentage \\ \hline
	Difference in Detail & A headline conveys additional details, such as a name, or a cause & \textit{NASA delays work on Moon rocket during virus pandemic \newline Nasa's Moon plans take a hit} & 37\% \\ \hline
	Exact Paraphrase & Both headlines convey the same information & \textit{Equifax takes web page offline after reports of new cyber attack\newline Equifax takes down web page after reports of new hack} & 30\% \\ \hline
	Difference in Focus & Headlines focus on a different aspect of the event group & \textit{Astronauts to Get Thanksgiving Feast in Space\newline A Brief History of Thanksgiving Turkey in Space} & 26\% \\ \hline
	Pun, Play-on-word, etc. & A headline has a unique stylistic element to attract readers & \textit{New privacy law forces some U.S. media offline in Europe\newline US websites blacked out in Europe on 'Happy GDPR Day'} & 7\% \\ \hline \hline
	\multicolumn{4}{|c|}{\textbf{Negative Examples}} \\
	Reasoning & Description & Example Headline Pair & Percentage \\ \hline
	Independent events & Headlines describe two distinct events involving common actors & \textit{Brazil dam disaster leaves 34 dead, hundreds missing \newline Alert raised over imminent risk to another Brazil mining dam} & 44\% \\ \hline
	Related Sub-events & Describing two events that are related, e.g., follow each other & \textit{Irish abortion referendum voting opens\newline Ireland set to end abortion ban as exit polls signal landslide vote} & 36\% \\ \hline
	Headline too broad & One of the headlines is too broad to assign to a particular event & \textit{Obama commutes Chelsea Manning sentence\newline Who is Chelsea Manning? - video profile} & 16\%                \\ \hline
	Borderline / Noise & Could be positive or negative, based on interpretation & & 4\% \\ \hline
    \end{tabular}
    }
    \caption{\textbf{Results of a manual typology of a subset of HLGD}. An analysis of 200 positive, and 200 negative same-day headline pairs reveals there are several reasons why headlines get grouped or not in our dataset.}
    \label{table:positive_negative}
\end{table*}

\begin{figure}
    \centering
    \includegraphics[width=0.47\textwidth]{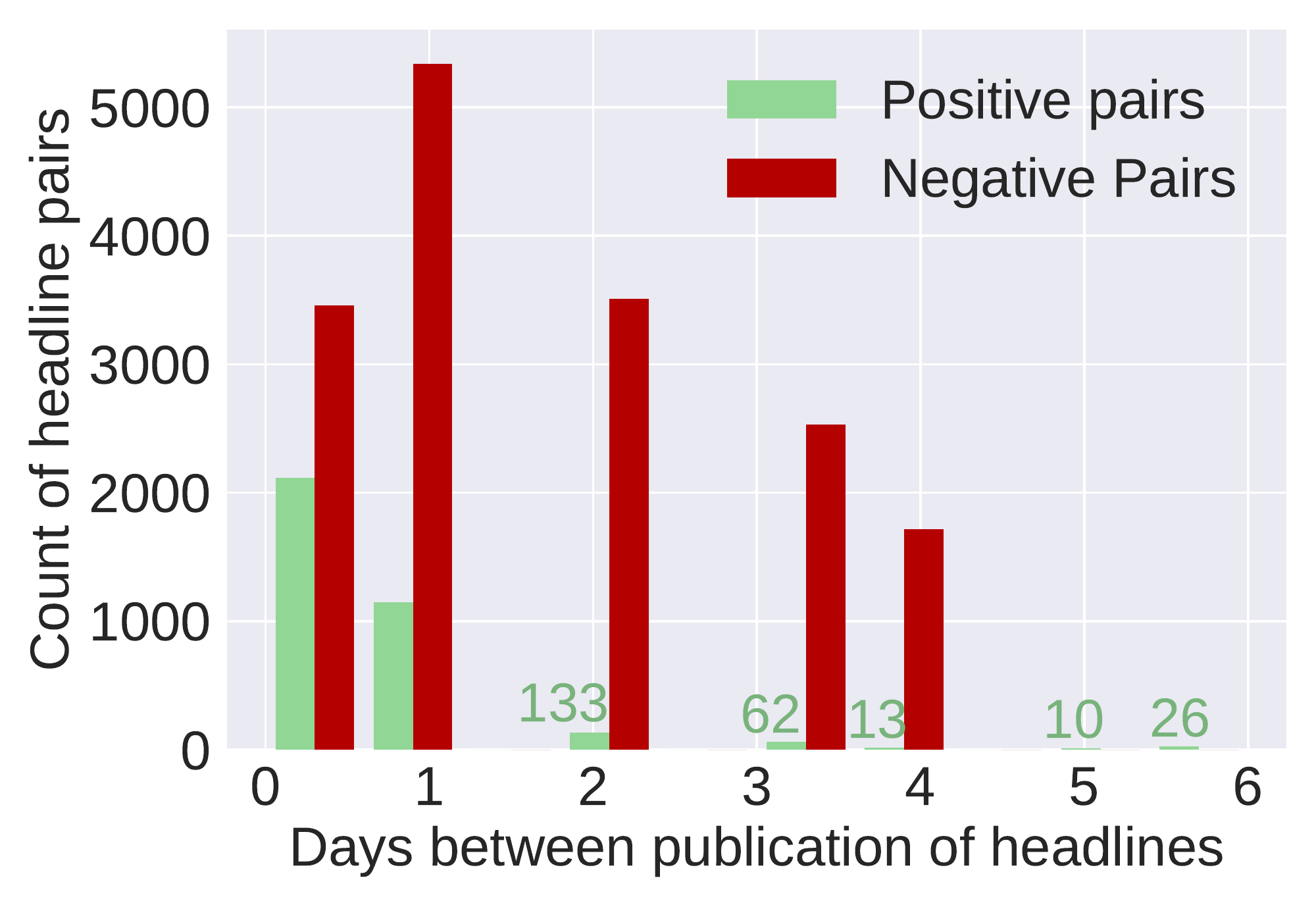}
    \caption{\textbf{Histogram of difference in publication dates of positive and negative pairs in final dataset.} Most positive headline pairs are published on the same day or within one day of each other. We down-sample negative pairs to only keep headlines published within 4 days of each other.}
    \label{fig:time_analysis}
\end{figure}

We transform the global groups into a binary classification task by generating pairs of headlines in the timelines: labeling the pair with a 1 if it belongs to the same group, and 0 otherwise.

With this procedure, we obtain 3,522 distinct positive headline pairs, and 154,156 negative pairs. This class imbalance is expected: two headlines picked at random in a timeline are unlikely to be in the same group. In order to reduce class imbalance, we down-sample negative pairs in the dataset.

Figure~\ref{fig:time_analysis} shows the distribution of differences in publication dates for pairs of headlines in the final dataset. Publication date is indeed a strong signal to determine whether headlines are in the same group, as most positive pairs are published on the same day or one day apart. However, we show in Section~\ref{section:results} that using time as a sole indicator is not enough to perform well on the dataset.

In Figure~\ref{fig:time_analysis}, it can also be observed that 98$\%$ of positive headline pairs are published within 4 days of each other. Therefore, we only retain negative pairs that are within a 4 day window, filtering out simpler negative pairs from the final dataset. This final dataset has a class imbalance of roughly 1 positive pair to 5 negative pairs, for a total of 20,056 of labeled headlines pairs. This is similar in size to other NLU datasets, such as MRPC (5,801 samples), or STS-B (8,628 samples). 

\begin{figure}
    \centering
    \includegraphics[width=0.47\textwidth]{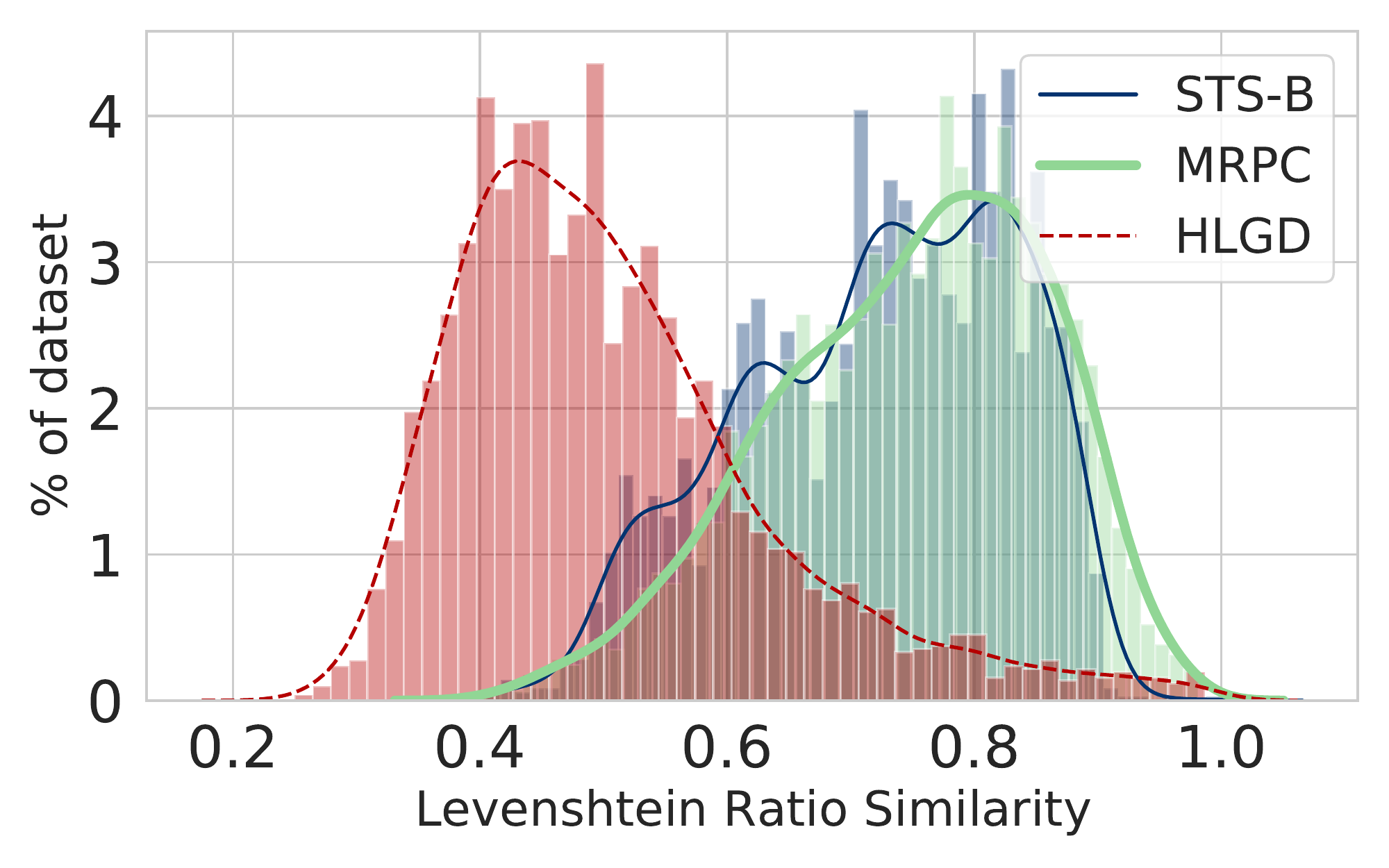}
    \caption{\textbf{Levenshtein Similarity Distribution for positive pairs in  MRPC, SST-B and HLGD.} 
    Pairs of sentences in HLGD can be positive examples (i.e., in the same group) while being less syntactically similar than in other NLU datasets such as STS-B or MRPC.}
    \label{fig:levenshtein_distance}
\end{figure}

Figure~\ref{fig:levenshtein_distance} shows the distribution of Levenshtein Ratio \cite{Levenshtein1966BinaryCC} defined as:
\begin{equation}
    Ratio(S_1, S_2) = 1 - \frac{Levenshtein(S_1, S_2)}{max(\vert S_1 \vert, \vert S_2 \vert)}
\end{equation}
for positive pairs $(S_1, S_2)$ in MRPC and STS-B, two common NLU datasets, as well as HLGD, computed at the character level. The average similarity in HLGD (0.51) is lower than in the two others (0.72 and 0.74, respectively). Furthermore, a classifier using solely the Levenshtein Ratio obtains an F-1 score of 0.81 on MRPC, but only 0.485 on HLGD. This suggests lexical distance alone does not contain a strong signal for good performance on HLGD.

\subsection{Analysis}

To gain insight into the linguistic phenomena that occur within and outside headline groups, the first author manually inspected 200 positive and 200 negative headline pairs in HLGD. Positive pairs were selected from randomly sampled large groups, and negative samples from same-day negative pairs, because headlines that appear on the same day but are not in the same group cannot be distinguished using time information and are likely to overlap semantically the most. In Table~\ref{table:positive_negative}, we list the phenomena we observed, give an example for each, and show the frequency in our sample.
Within a group, headlines can be exact paraphrases, differ in detail level, differ in the element of focus, or involve stylistic elements such as puns.
Negative headline pairs analyzed were either about independent events, related sub-events or involved a headline that was not specific enough. Additionally, around 4$\%$ of the negative samples analyzed were judged as borderline, interpretable as either positive or negative, showing that some ambiguity in the task is unavoidable.
We believe this diversity in phenomena are ingredients that make HeadLine Grouping challenging and interesting for NLU research.

\subsection{Challenges}

To allow for diversity in approaches to HeadLine Grouping, we propose to sub-divide HLGD into several challenges, limiting in each the data used to solve the classification task:

\begin{itemize}
    \item \textbf{Challenge 1: Headline-only}. Access to the headline pairs only; similar to Paraphrase Identification and Textual Similarity tasks.
    \item \textbf{Challenge 2: Headline + Time}.
    Access to the headline pairs and their publication dates.
    \item \textbf{Challenge 3: Headline + Time + Other}. Access to the headline pairs, publication dates, and other information such as full content, author(s), and news source (a URL to the original article provides this access).
\end{itemize}

We believe these different challenges provide flexibility to probe a diversity of methods on the HLGD task. Challenge 1 fits the standard text-pair classification of NLU, similar to paraphrase identification, textual similarity and NLI, while additional meta-data available in Challenge 3 might be more compatible with the goals of the information retrieval community.

\section{Results on the Challenges}
\label{section:results}

\begin{table*}[]
    \centering
    \resizebox{0.87\textwidth}{!}{%
	\begin{tabular}{lccc}
	\hline
	\textbf{Method} & \textbf{Challenge} & \textbf{HLGD Dev F-1} & \textbf{HLGD Test F-1} \\ \hline
	\multicolumn{4}{c}{\textbf{Human Performance and Baseline}} \\ \hline
	Human Performance & 3 & \textbf{0.884} & \textbf{0.900} \\
	Time-only & 2 & 0.654 & 0.585 \\ \hline
	\multicolumn{4}{c}{\textbf{Unsupervised / Zero-shot Models}} \\ \hline
	Electra MRPC Zero-Shot & 1 & 0.562 & 0.626 \\
	Electra MRPC Zero-Shot + Time & 2 & 0.666 & 0.688 \\
	Headline Generator Swap & 3 & 0.671 & 0.651 \\
	Headline Gen. Swap + Time & 3 & \textbf{0.727} & \textbf{0.722} \\ \hline
	\multicolumn{4}{c}{\textbf{Supervised Models}} \\ \hline
	Electra Finetune on HLGD & 1 & 0.728 & 0.796 \\
	Electra Finetune on HLGD content & 3 & 0.652 & 0.723 \\
	Electra Finetune on HLGD + Time & 2 & \textbf{0.753} & \textbf{0.828} \\ \hline
	\end{tabular}
    }
    \caption{\textbf{F-1 performance of several methods on the development and test portions of the HeadLine Grouping Dataset.} Methods are separated into baselines: (1) the performance of a human annotator, and performance using only publication date, (2) unsupervised or zero-shot methods that do not leverage the training set for predictions, and (3) supervised methods. Each method falls under a challenge setting (1, 2 or 3) based on data used.}
    \label{table:results}
\end{table*}

In Table~\ref{table:results}, we report the performance of a human annotator and a baseline, as well as unsupervised and supervised methods on HLGD. We chose Electra \cite{Clark2020ELECTRA} for experiments based on a bi-directional Transformer \cite{vaswani2017attention}, as initial experiments with other BERT \cite{devlin2019bert} variants performed similarly. Implementation details, model sizes and hyper-parameters are listed in Appendix~\ref{appendix:model_details}.

\begin{figure}
    \centering
    \includegraphics[width=0.47\textwidth]{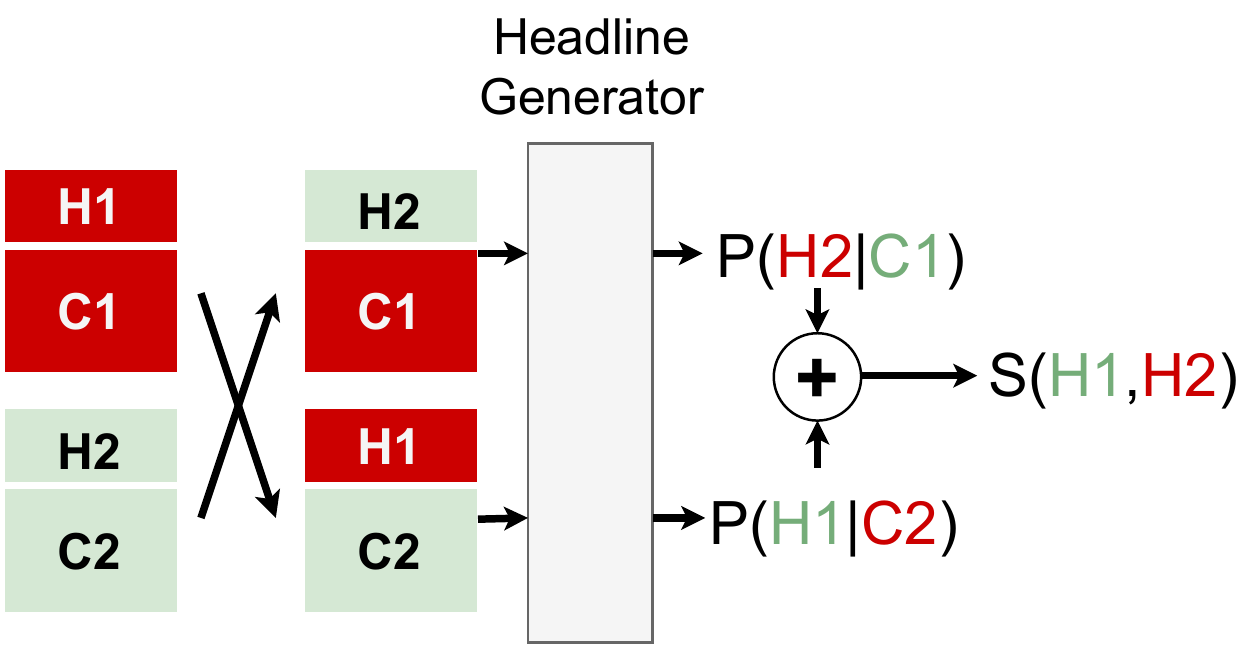}
    \caption{\textbf{Schematic of the Headline Generator Swap model.} We adapt a headline generator -- a finetuned GPT-2 -- to the task of HeadLine Grouping. The score of a pair of headlines is based on whether the generative model believes a swap of headlines is likely.}
    \label{fig:headline_generator_swap}
\end{figure}

\subsection{Human Performance and Baseline}
\textbf{Human Performance} reports the F-1 score of human annotators performing the task. Human performance is estimated by obtaining a sixth set of annotations for each timeline in the development and testing set, beyond the five used for dataset creation. These annotations were completed after several hours of practice on the training set timelines.

Human performance is distinct from the inter-annotator agreement (IAA) analysis presented in \S\ref{sec:inter_annotator_agreement}. IAA was performed on the five annotations used to create the dataset. We note that human performance can theoretically achieve a perfect F-1 score of 1.0 if the sixth annotator grouped the headlines identically to the global group.

\textbf{Time only} reports the performance of a logistic regression baseline based on the difference in days of publication between the two headlines. Data plotted in Figure~\ref{fig:time_analysis} shows that a majority of positive pairs are published within two days of each other.

\subsection{Unsupervised Models}

\textbf{Electra MRPC Zero-shot} stands for an Electra model trained on the Microsoft Paraphrase Corpus (MRPC), achieving an F-1 of 0.92 on its development set. The objective is to evaluate whether a competitive paraphrase identification system achieves high performance on HLGD. The threshold to predict a label of one is tuned on the training portion of HLGD. This model only accesses headlines, and falls under \textit{Challenge 1}.

\textbf{Electra MRPC Zero-shot + Time} corresponds to the previous model, adding publication time into the model in the following way:
\begin{equation}
    P'(Y=1 \vert X) = P(Y=1 \vert X) \cdot e^{-\lambda  \Delta T}
\end{equation}

\noindent
where $X$ represents the pair of headline inputs, $P(Y=1 \vert X)$ represents the model's confidence of the headline pair being in the same group, and $\Delta T$ the difference in days of publication of the headlines. $\lambda$ is tuned on the training set.
Because this method leverages headline and time information, it falls under \textit{Challenge 2}.

\textbf{Headline Generator Swap} is a novel approach we propose for zero-shot headline grouping, summarized in Figure~\ref{fig:headline_generator_swap}.

Transformer-based Language Models like GPT-2 \cite{radford2019language} model the probability of a word sequence. As the first step in Headline Generator Swap, we use the GPT-2 model to create a headline generator to estimate the likelihood of a (headline, content) pair: $P_{LM}(H \vert C)$.

In more detail, we finetune a GPT-2 model to read through the first 512 words of a news article and generate its headline. The headline generator is trained with teacher-forcing supervision, and a large corpus of 6 million (content, headline) pairs \cite{laban2017newslens}, not overlapping HLGD.

The second step in Headline Generator swap is to use this probability to produce a symmetric score for two articles $A_1 = (H_1, C_1)$ and $A_2 = (H_2, C_2)$:
\begin{equation}
    S(A_1, A_2) = P_{LM}(H_2 \vert C_1) + P_{LM}(H_1 \vert C_2)
\end{equation}
This score evaluates the likelihood of a swap of headlines between articles $A_1$ and $A_2$, according to the GPT-2 language model. We argue that if the model believes a swap is likely, the headlines must be part of the same group.
The threshold above which $S(A_1, A_2)$ predicts a 1 is determined using the training portion of the data.
Because this model uses the headline and content of the article, it falls under \textit{Challenge 3}. 

\textbf{Headline Gen. Swap + Time} corresponds to the Headline Generator Swap model, adding publication date information similarly to the \textit{Electra MRPC Zero-shot + Time} model:
\begin{equation}
    S'(A_1, A_2) = S(A_1,A_2) \cdot e^{-\lambda  \Delta T}
\end{equation}
This model uses the headline, publication data and content of the article, and falls under \textit{Challenge 3}.

Unsupervised models were allowed to pick a single hyper-parameter based on training set performance: to learn the threshold in score differentiating between class 1 and class 0. Strictly speaking, because we tune this single parameter, the methods could be seen as supervised. However, we label them as unsupervised because model parameters were not modified.

\subsection{Supervised Methods}

\textbf{Electra Finetune} stands for an Electra model finetuned on the training set of HLGD, inputting the two headlines, divided by a separator token. Headline order is chosen randomly at each epoch. Because we train a model for several epochs (see Appendix~\ref{appendix:model_details}), a model is likely to see pairs in both orders. This model only uses headlines of articles for prediction, and falls under \textit{Challenge 1}.

\textbf{Electra Finetune on content} represents a similar model to that described above, with the difference that the model makes predictions based on the first 255 words of the contents of the two news articles, instead of the headline. This evaluates the informativeness of contents in determining headline groups. This experiment requires the contents and falls under \textit{Challenge 3}.

\textbf{Electra Finetune + Time} corresponds to an Electra model with time information. The model's output goes through a $768$x$1$ feed-forward layer, and is concatenated with the day difference of publication, which is run through a $2$x$2$ feed-forward, and a \texttt{softmax} layer. This model uses headline and time information, and falls under \textit{Challenge 2}.

\subsection{Discussion of Results}

Human performance can be high, close to 0.9 F-1 both on development and test timelines. 

Using time alone gives a lower-bound baseline on HLGD, achieving an F-1 of 0.585 on the test set, and confirming that publication date of an article is not enough to perform competitively on HLGD.

Regarding Unsupervised and Zero-shot approaches, the \textit{Headliner Generator Swap} outperforms \textit{Electra MRPC Zero-shot}. With additional time information (\textit{+ time}), the generator-based model is able to get close to strong supervised models. The model benefits from pre-training on a large corpus of (content, headline) pairs, having learned a good representation for headlines.

Unsurprisingly, best performance on HLGD is achieved by a supervised approach, \textit{Electra Finetune HLGD + Time}, which uses both headline and time information. With an F-1 performance on the development set of 0.753, the model is still 0.13 F-1 points below human performance (0.07 F-1 difference on the test set).

When finetuning the Electra model with contents instead of headlines, performance drops by 0.07 F-1 points. This is particularly surprising as it could be expected that content contains strictly more information than the headline. We interpret this performance of the content-based model as evidence that the contents are more broad and do not solely focus on the distinguishing fact that is necessary to perform the grouping.

Finally, publication date yields a performance gain of 0.025 to 0.1 F-1 points over models without time information. This confirms that even though time information alone does not achieve high performance, it can be used to enhance models effectively. Because human annotators read timelines chronologically and had access to publication date while annotating, we do not have an upper-bound of human performance without using time.

\section{Analysis of Model Consistency}
\label{section:analysis}

\begin{figure}
    \centering
    \includegraphics[width=0.43\textwidth]{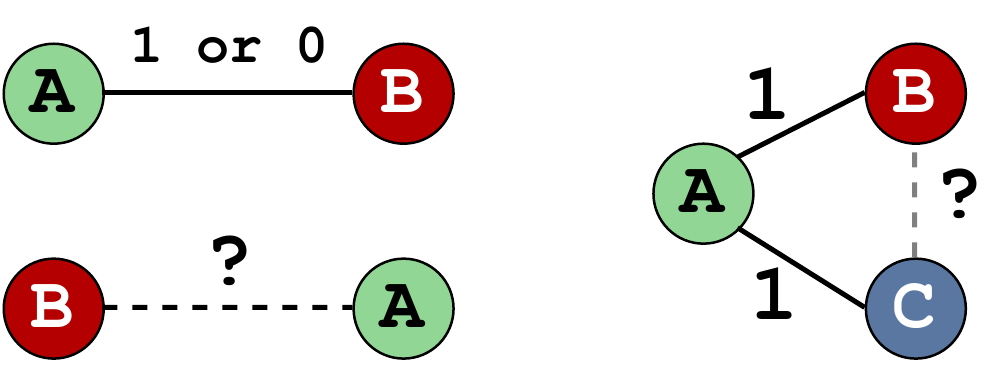}
    \caption{\textbf{Simplified representation of commutativity (left) and transitivity (right).} We verify whether predictions from our best-performing model are consistent with regards to these two properties.}
    \label{fig:consistency_diagram}
\end{figure}

Checking whether deep learning models are consistent across predictions has recently become a subject of interest, for example with QA systems with text \cite{Ribeiro2019AreRR} and image \cite{shah2019cycle} inputs.
We analyze model consistency by probing the \textit{Electra Finetune + Time} model, which achieves highest performance in terms of F-1 score. We propose a commutative test and transitive test, both illustrated in Figure~\ref{fig:consistency_diagram}.

In order to evaluate consistency across training runs, we trained six versions of the \textit{Electra Finetune + Time} model with the same hyper-parameters. Because each training run processes through the data in a different order, the models are distinct from each other. With regard to performance, the models perform very similarly, achieving within 0.01 F-1 of each other on the development and test sets.

\subsection{Commutative Test}

The HeadLine Grouping task requires two sentences to be compared, both playing a symmetric role.

Most model architectures process the headline pair as a single sequence, and an arbitrary ordering of the pair is chosen for processing.
We study whether this arbitrary choice has an impact on the model's prediction. Specifically, we make predictions for all pairs of headlines in the development portion of HLGD, running each pair in both $(A,B)$ and $(B,A)$ order.

On average across the 6 model checkpoints, swapping the order of headlines is enough to make the model change its prediction (put higher probability on 0 in one case and 1 in the other) on $6.3\% (\pm 0.5)$ of the pairs.

Furthermore, in other cases when the prediction does not change, the probability of the predicted class fluctuates by $0.061 (\pm 0.005)$ on average, showing the impact sentence order has on all predictions.

The relatively small standard deviations across training runs indicates that this phenomenon is inherent to the training procedure and not only existent in a subset of models.

A remedy is to  build a symmetric classifier:

\begin{equation}
    P_{S}(Y \vert A, B) = \frac{P(Y | A,B) + P(Y | B, A)}{2} 
\end{equation}

\noindent
where $P_S$ follows the symmetric rule by design, by predicting for both $(H_1,H_2)$ and $(H_2, H_1)$ and averaging.
When applying this patch to models presented in Section~\ref{section:results}, we observe an average gain in F-1 performance of 0.01. Even though encouraging, this gain is a post hoc fix, and enforcing symmetry during training might yield further gains.

\subsection{Transitive Test}

Transitivity involves triplets of headlines A, B and C. The assumption is that if A and B are part of the same group, and A and C are part of the same group, then B and C must be in the same group as well.
The procedure followed during annotation \textemdash~assigning group IDs to headlines \textemdash~implies that the transitivity is preserved, as all headline pairs within the same group are positive pairs.

To test a model's consistency with regards to the transitive rule, we use the \textit{Electra Finetune + Time} model to produce a prediction for all pairs of headlines in the development portion of HLGD.

For each triplet (A,B,C) of headlines in the timeline, the model produces three predictions for the (A,B), (A,C), and (B,C) pairs. We focus our attention on triplets where the model has predicted at least 2 positive pairs: if the third pair is predicted to be positive, transitivity is conserved (111 triangle), but if it is predicted to be negative, the triplet breaks the transitivity rule (110 triangle).

On average across the six model checkpoints, we find that of the 60,660 triplets for which the model predicted at least 2 positives pairs, 44,627 triplets had a negative third prediction, and 16,033 had a positive one. In short, the model is consistent only $26.4\% (\pm 1.4)$ of the time on these triplets.

Improving model consistency with regards to transitivity is challenging, as it would involve presenting the model with triples in some way. Imposing this constraint could yield performance improvements on the task.

We note however that transitivity is a strong assumption, as it is possible for groups of headlines to have stronger and weaker subgroups. It is possible that human annotations would not always follow transitivity if tasked to do so. For this reason, we do not expect models to be $100\%$ consistent, but there is room for improvement.

\section{Conclusion}

In this work we present the new task of HeadLine Grouping (HLG) a new challenging NLU task, with an accompanying dataset (HLGD). Even though state-of-the-art NLU models have achieved close to human performance on many NLU tasks, we show that there is a considerable gap between best model performance (0.75 F-1) and human performance (about 0.9 F-1) on HLGD. We therefore propose this dataset as a challenge for future NLU benchmarks.
We propose to repurpose a Headline Generator for the task of headline grouping, based on prompting it for the likelihood of a headline swap, and achieve within 3 F-1 of the best supervised model, paving the way for other unsupervised methods to repurpose generators for NLU.
Analysis of models on HLGD reveals that they are not consistent in trivial ways, suggesting further improvements needed to NLU models.

\section*{Acknowledgments}

We would like to thank the Upwork crowd-workers for their assistance in creating HLGD, as well as Katie Stasaski, Dongyeop Kang and the ACL reviewers for their helpful comments. This work was supported by a Bloomberg Data Science grant. We also gratefully acknowledge support received from an Amazon Web Services Machine Learning Research Award and an NVIDIA Corporation GPU grant.

\bibliography{naacl2021}
\bibliographystyle{acl_natbib}

\appendix
\renewcommand{\thetable}{A\arabic{table}}
\setcounter{table}{0}
\renewcommand{\thefigure}{A\arabic{figure}}
\setcounter{figure}{0}

\section{Annotator Instructions}
\label{appendix:annotator_intructions}
The wording of the prompt given to the eight crowd annotators we recruited was the following:
\begin{quote}
    Your task will be to annotate News Headline timelines, and decide which are referring to the same event.
    
    You are given a list of news headlines in chronological order, with a headline on each line of a Spreadsheet. For each headline, the task is to assign it a number: either a new number if the headline represents a new event that hasn't appeared yet, or the number of the existing headline it is a ``repetition'' of.
    
    For each headline, you are also given a date of publication, which you can use to determine whether two headlines should be in the same event as well: two headlines several months apart must be about different events, even if they are very close lexically (protests in June and September are different events).
\end{quote}

In some cases, headlines in the timelines were too vague or did not describe an event specifically, and annotators were encouraged to put such headlines in a group of their own.

\section{Model Size and Hyper-parameters}
\label{appendix:model_details}
In order to ease reproducibility, we report relevant hyper-parameters of models whose results are present in Section~\ref{section:results}. We used implementation of Transformer models from the HuggingFace Transformer library\footnote{https://github.com/huggingface/transformers}. For Electra models, we initialized using the \textit{electra-base-discriminator}. For GPT-2 based models, we initialized with the \textit{gpt2} model (which corresponds to a base model as well). Additional model-specific details:

\begin{itemize}
    \item \textbf{Electra MRPC Zero-shot:} The model produces a probability for label 1: $P(Y=1 \vert X)$. If this probability is above a threshold $T$, the model predicts a 1, and below it predicts a 0. $T = 0.23$ for this model.
    
    \item \textbf{Electra MRPC Zero-shot + Time:} We use a time constant of $\lambda = 0.15$, and $T = 0.14$ for this model.
    
    \item \textbf{Headline Generator Swap:} the threshold for this model is $T = 0.0012$. This might seem small, but it is the conditional probability of a headline according to the $GPT-2$ model, and corresponds to a log-probability of $-6.75$.
    
    \item \textbf{Headline Gen. Swap + Time:} we use a time constant $\lambda = 0.07$, and a threshold of $T = 0.00056$, which corresponds to a log-probability of $-7.49$.
    
    \item \textbf{Supervised models:} All supervised models are trained for 3 epochs, with a batch size of 32, and the Adam Optimizer with a learning rate of $LR = 10^{-5}$ with an exponential decay and a linear-warmup over the first 1000 steps. All weights of the model are finetuned.
\end{itemize}

\section{HLGD Format and Removal process}
\label{appendix:dataset_format}
The dataset is a JSON file that can be processed using standard JSON parsing libraries. Each entry in the JSON object follows the schema:
\begin{verbatim}
    {
    "headline_a": "...",
    "headline_b": "...",
    "day_a": "YYYY-MM-DD",
    "day_b": "YYYY-MM-DD",
    "source_a": "domain.com",
    "source_b": "domain.com",
    "authors_a": "...",
    "authors_b": "...",
    "url_a": "https://...",
    "url_b": "https://...",
    "cut": "...",
    //training/validation/testing
    "timeline": "",
    // name of timeline
    // headline pair belongs to
    "label": int,
    // 1 if paraphrase, 0 o/w 
    }
\end{verbatim}

The dataset will include scripts for processing the url for accessing the full content of the articles and other article data and for an option for content owners to request removal.

\section{Excerpt of Timeline}
\label{appendix:longer_excerpt}
\begin{table*}[]

\resizebox{0.96\textwidth}{!}{%
\begin{tabular}{llp{11cm}c}
    \hline
    \begin{tabular}[c]{@{}l@{}}\textbf{Publication}\\ \textbf{ Date}\end{tabular} & \textbf{Source}             & \textbf{Headline}                                                                                   & \textbf{Group} \\ \hline
    2015-01-14                                                 & cnn            & Astronauts relocate after false alarm                                                      & 1     \\
    2015-01-14                                                 & bloomberg      & Space Station Crew Returns After Alarm Scare Prompts Evacuation                            & 1     \\
    2015-01-14                                                 & bloomberg      & Space Station Crew Safe After Coolant-Pressure Alarm Sounds (1)                            & 1     \\
    2015-01-14                                                 & foxnews        & 6 evacuate US part of space station; NASA says all are safe                                & 1     \\
    2015-01-15                                                 & reuters        & Astronauts back in U.S. part of space station after leak scare                             & 1     \\
    2015-01-15                                                 & reuters        & Crew evacuates U.S. section of space station after leak-agencies                           & 1     \\
    2015-01-15                                                 & nytimes        & Space Station Crew Temporarily Moves to Russian Side Over Fears of Ammonia Leak            & 1     \\
    2015-01-17                                                 & washingtonpost & A false alarm for crew on the International Space Station                                  & 1     \\
    2015-08-10                                                 & telegraph    & Astronauts declare first space salad 'awesome'                                             & 2     \\
    2015-08-10                                                 & cnn            & Space-grown vegetables: Astronauts chow down on lettuce                                    & 2     \\
    2015-08-10                                                 & foxnews        & For the First Time Ever, NASA Astronauts Eat Vegetables Grown in Space                     & 2     \\
    2015-08-10                                                 & foxnews        & Space Station astronauts make history, eat first space-grown veggies                       & 2     \\
    2015-08-10                                                 & businessinsider & First space-grown lettuce on the menu today for NASA astronauts                            & 2     \\
    2015-08-11                                                 & nytimes        & Growing Vegetables in Space, NASA Astronauts Tweet Their Lunch                             & 2     \\
    2016-11-16                                                 & ap             & NASA astronaut on verge of becoming oldest woman in space                                  & 3     \\
    2016-11-16                                                 & washingtonpost & Astronaut to become oldest woman to travel in space                                        & 3     \\
    2016-11-17                                                 & france24       & Haute cuisine: top French chefs' food bound for space station                              & 4     \\
    2016-11-17                                                 & ap             & Rocket carrying crew of 3 blasts off for Int'l Space Station                               & 5     \\
    2016-11-17                                                 & reuters        & Multinational crew blasts off, bound for space station                                     & 5     \\
    2016-11-17                                                 & rt             & New ISS crew sets off into space from Russian launchpad (LIVE)                             & 5     \\
    2016-11-17                                                 & france24       & Three astronauts blast off to ISS                                                          & 5     \\
    2016-11-17                                                 & foxnews        & Rocket carrying crew of 3 blasts off for International Space Station                       & 5     \\
    2016-11-17                                                 & bbc          & Peggy Whitson: Oldest woman in space blasts off to ISS                                     & 5     \\
    2016-11-18                                                 & telegraph    & Nasa veteran Peggy Whitson becomes the oldest woman in space as she arrives at the ISS     & 5     \\
    2016-11-18                                                 & theguardian    & Oldest woman in space blasts off again for third ISS mission                               & 5     \\
    2016-11-18                                                 & bbc          & Peggy Whitson: Blast off to the ISS for oldest woman in space                              & 5     \\
    2016-11-19                                                 & telegraph    & Russian spaceship delivers three astronauts to space station                               & 5     \\
    2016-11-19                                                 & ap             & Space station receives oldest female astronaut, bit of Mars                                & 5     \\
    2016-11-20                                                 & foxnews        & Space station welcomes the oldest woman astronaut, and a bit of Mars                       & 5     \\
    2016-11-20                                                 & france24       & Space station welcomes Frenchman and world's oldest astronaut                              & 5    \\
    2016-11-23                                                 & bbc          & Waste not, want not                                                                        & 6     \\
    2016-11-24                                                 & france24       & French astronaut Pesquet describes first days aboard space station                         & 7     \\
    2016-11-25                                                 & telegraph    & French astronaut lands on International Space Station - and is asked to fix the loo        & 7     \\
    2017-05-12                                                 & bloomberg      & NASA Rejects Idea of Humans on First Flight of New Rocket                                  & 8     \\
    2017-05-12                                                 & bloomberg      & NASA Study Warns Against Putting Crew On Huge Rocket's First Flight                        & 8     \\
    2017-05-13                                                 & reuters        & NASA delays debut launch of \$23 billion moon rocket and capsule                           & 9     \\
    2018-10-03                                                 & france24       & NASA skeptical on sabotage theory after mystery ISS leak                                   & 10    \\
    2018-10-03                                                 & theguardian    & Nasa casts doubt on Russian theory ISS air leak was sabotage                               & 10    \\
    2018-10-03                                                 & independent  & Nasa casts doubt on claims International Space Station leak was deliberate                 & 10    \\
    2018-10-04                                                 & france24       & Astronauts return to Earth from ISS amid US-Russia tensions                                & 11    \\
    2018-10-04                                                 & france24       & ISS astronauts return to Earth amid US-Russia tensions                                     & 11    \\
    2019-04-11                                                 & nytimes        & Scott Kelly Spent a Year in Orbit. His Body Is Not Quite the Same.                         & 12    \\
    2019-04-11                                                 & cnn            & Human health can be 'mostly sustained' for a year in space, NASA Twins Study concludes     & 12    \\
    2019-04-11                                                 & bloomberg      & NASA's Twins Study Sees No Red Flags for Human Space Travel                                & 12    \\
    2019-04-11                                                 & independent  & Space is doing strange things to astronauts' bodies, Nasa study reveals                    & 12    \\
    2019-04-11                                                 & reuters        & Oh, brother! NASA twins study shows how space changes the human body                       & 12    \\
    2019-04-11                                                 & france24       & NASA's 'Twins Study,' landmark research for an eventual Mars mission                       & 12   \\ \hline
    \end{tabular}
    }
    \caption{\textbf{Excerpt of the timeline about the International Space Station in HLGD.} Group is the global group, aggregating labels from the five annotators. The full timeline contains 257 headlines and 107 distinct groups.}
    \label{table:space_timeline}

\end{table*}

\end{document}